\documentclass[journal,twoside,web]{ieeecolor}
\usepackage{tmi}

\usepackage{amsmath,amssymb,amsfonts}
\usepackage{algorithm}
\usepackage{algpseudocode}

\usepackage{array}
\usepackage{utfsym}
\usepackage{booktabs}
\usepackage[caption=false,font=normalsize,labelfont=sf,textfont=sf]{subfig}
\usepackage{pifont}
\usepackage{makecell}
\usepackage{diagbox}
\usepackage{titlesec}
\usepackage{stfloats}
\usepackage{multirow}
\usepackage{verbatim}
\usepackage{graphicx}
\usepackage{textcomp}
\usepackage{orcidlink}
\usepackage{cite}
\usepackage{url}
\usepackage{color}
\makeatletter
\let\NAT@parse\undefined
\makeatother
\usepackage{hyperref}
\hypersetup{
        hypertex=true,
	colorlinks=true,
	linkcolor=blue,
	filecolor=blue,      
	urlcolor=black,
	citecolor=blue,
}

\def\BibTeX{{\rm B\kern-.05em{\sc i\kern-.025em b}\kern-.08em
    T\kern-.1667em\lower.7ex\hbox{E}\kern-.125emX}}
\begin{document}
\title{MacNet: An End-to-End Manifold-Constrained Adaptive Clustering Network for Interpretable Whole Slide Image Classification}

\author{Mingrui Ma*~\orcidlink{0000-0002-4851-4356}, Chentao Li*~\orcidlink{0009-0001-5474-9311}, \IEEEmembership{Graduate Student Member, IEEE}, Pan Huang~\orcidlink{0000-0001-8158-2628}, \IEEEmembership{Member, IEEE} and Jing Qin~\orcidlink{0000-0002-7059-0929}, \IEEEmembership{Senior Member, IEEE}
\thanks{\textit{Corresponding author: Pan Huang.}}
\thanks{Mingrui Ma is with the School of Computer Science and Technology, Xinjiang University, Urumqi 830046, China, and the Affiliated Tumor Hospital of Xinjiang Medical University, Xinjiang 830011, China (e-mail: mamingrui@stu.xju.edu.cn).}
\thanks{Chentao Li is with the Fu Foundation School of Engineering and Applied Science, Columbia University, New York, NY 10027, USA (e-mail: cl4691@columbia.edu).}
\thanks{Pan Huang is with the School of Nursing, The Hong Kong Polytechnic University, Hong Kong 999077, SAR China (e-mail: mrhuangpan@163.com).}
\thanks{Jing Qin is with the School of Nursing, The Hong Kong Polytechnic University, Hong Kong 999077, SAR China (e-mail:harry.qin@polyu.edu.hk).}
\thanks{* Equal contributions.}}
\maketitle

\begin{abstract}
Whole slide images (WSIs) are the gold standard for pathological diagnosis and sub-typing. Current main-stream two-step frameworks employ offline feature encoders trained without domain-specific knowledge. Among them, attention-based multiple instance learning (MIL) methods are outcome-oriented and offer limited interpretability. Clustering-based approaches can provide explainable decision-making process but suffer from high dimension features and semantically ambiguous centroids. To this end, we propose an end-to-end MIL framework that integrates Grassmann re-embedding and manifold adaptive clustering, where the manifold geometric structure facilitates robust clustering results. Furthermore, we design a prior knowledge guiding proxy instance labeling and aggregation strategy to approximate patch labels and focus on pathologically relevant tumor regions. Experiments on multicentre WSI datasets demonstrate that: 1) our cluster-incorporated model achieves superior performance in both grading accuracy and interpretability; 2) end-to-end learning refines better feature representations and it requires acceptable computation resources. The code will be available at \href{https://github.com/Prince-Lee-PathAI/MacNet}{\textit{\textbf{https://github.com/Prince-Lee-PathAI/MacNet}}}.
\end{abstract}


\section{Introduction}\label{intro}
\label{sec:introduction}
\IEEEPARstart{A}{s} the gold standard for pathological grading, whole slide images (WSIs) play a crucial role in diagnosis, prognosis, and migration \cite{kumar2020whole}. Whereas, it is a time-consuming job to examine WSIs even for expert pathologists. Recently, multi-instance learning (MIL) has served as a promising paradigm for analyzing giga-pixel WSIs in computational pathology. It is also a weakly supervision problem where slide-level (bag) labels are typically accessible but enormous instance-level (patch) labels cropped from bags are unavailable.

\begin{figure*}[!t]
\centering
{\includegraphics[width=6in]{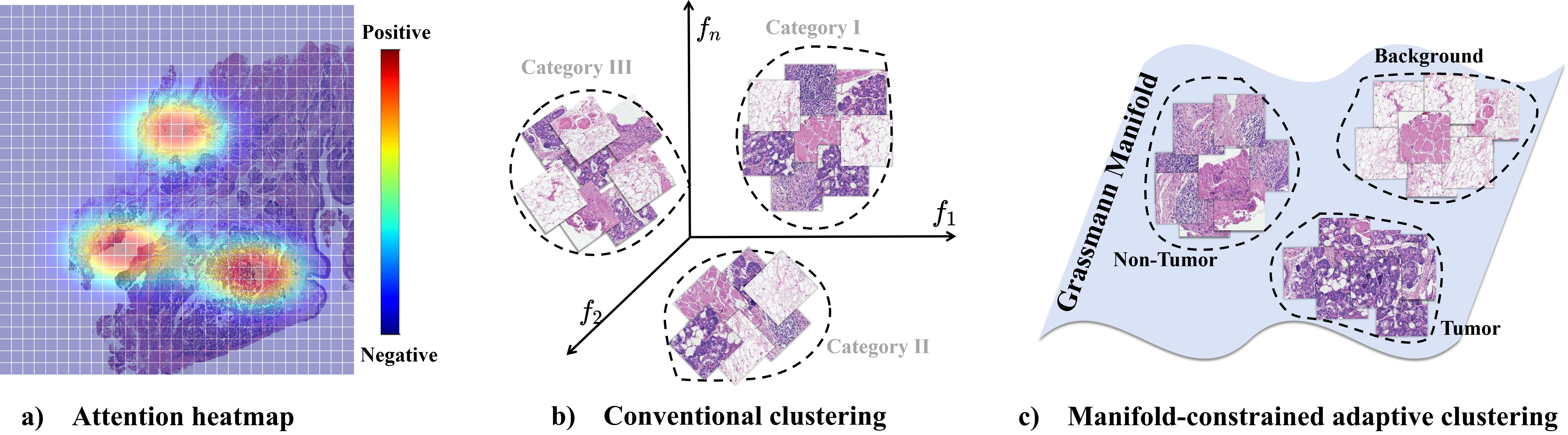}}
\hfil
\caption{Motivation of MacNet. \textbf{a):} Attention heatmap only reflects outcome-guided scores. \textbf{b):} Traditional clustering suffers from high dimension and ambiguous centroids. \textbf{c)}: MacNet obtains robust clustering results and transparent decision-making.}
\label{fig:fig1}
\end{figure*}

In this case, effectively modeling inter-instance relationships and aggregation mechanisms are critical to the performance of WSI classification. Attention-based models are widely adopted in MIL frameworks \cite{li2021dual, shao2021transmil, ilse2018abmil, zhang2024acmil}. Although these models can produce approximate pseudo label, the resulting attention scores are typically outcome-oriented. In Figure \ref{fig:fig1} a), it reveals the correlation between instances and bag predictions, but fails to explain how the instances affect pathology patterns. Consequently, these attention-based models offer limited interpretability regarding the underlying decision process.

From a genetic biology perspective, tumor regions exhibit a stronger association with pathological grading patterns compared to non-tumor regions \cite{renshaw2014diagnostic}. Grouping instances by different categories, cluster-incorporated framework can specifically offer more explainable motivation and contribute to classification decisions. Nevertheless, directly applying traditional clustering methods often leads to suboptimal results due to two main reasons. For one thing, by assigning equal weight to each dimension, distance metrics in high-dimensional Euclidean space lose discriminative power, and informative features are diluted by noise. For another, the cluster labels in unsupervised settings remain ambiguous, such as FuzzyMIL \cite{liu2025fuzzymil}, a deep clustering framework based on learnable fuzzy C-means to analyze WSIs. These cluster-based models with unrobust and fuzzy centroids tend to overemphasize low-effect non-tumor instances (NTIs) and irrelevant background instances (BGIs), thus degrading interpretability and classification performance; see Figure \ref{fig:fig1} b).

Moreover, given the overwhelming computational cost, most existing WSI-based MIL methods adopt a two-stage pipeline rather than end-to-end learning. Typically, they extract patch-level features using encoders pretrained on large datasets such as ImageNet \cite{russakovsky2015imagenet}. These features capture only general patterns from natural images, failing to represent pathology-specific structures such as tissue architecture and cellular contours. Such Frozen encoder prevents the model from adapting to domain-relevant distinctions even for pathological WSI pretrained encoder like PLIP \cite{huang2023plip}, thereby limiting its ability to differentiate tumor instances related to grading tasks.

Clearly, WSIs data are partially sampled from pathological tissues, which is naturally aligned with the manifold assumption \cite{belkin2006manifold} that these data lie on some low-dimensional manifolds embedded in high-dimensional spaces. Inspired by the benign clustering properties of manifolds \cite{belkin2006manifold} and transfer learning insights from the Geodesic Flow Kernel (GFK) approach \cite{gong2012geodesic}, we propose a prior instances guiding end-to-end learning network for pathological WSI classification. It consists of two modules that are Grassmann re-embedding method and manifold adaptive clustering method. To address the ambiguous cluster centroids in unsupervised settings, we further introduce a proxy instance labeling and aggregation strategy. Figure \ref{fig:fig1} c) shows that these methods combined enable the model to focus on tumor regions that are relevant to pathological grading more effectively, thereby enhancing classification performance, interpretability, and decision transparency. Extensive experiments on two private and two public WSI datasets illustrate that through end-to-end optimization, our method outperforms state-of-the-art (SOTA) approaches. The contributions of proposed framework can be concluded as follows:

\begin{itemize}
    \item \textbf{Grassmann re-embedding}: We utilized the Geodesic Flow Kernel (GFK) method to re-embed the prior knowledge and unlabelled instances into Grassmann manifold subspace. Meanwhile, we innovated the original derivation and implementation for end-to-end learning.
    \item \textbf{Manifold adaptive clustering}: We proposed a novel manifold-constrained adaptive clustering method based on the geometric properties and metric learning, where instance in the manifold subspaces with less redundancy and more feature  can be easily and robustly clustered.
    \item \textbf{Proxy instance labeling and aggregation}: We developed a proxy-based instance labeling and aggregation strategy. We effectively address the ambiguity of cluster labels in a more interpretable manner. Through adaptively assigning different weights, the model tends to focus on most representative and pathologically influential instances.
\end{itemize}

\section{Realted Work}
\subsection{Two-step Multiple Instance Learning}
Main stream MIL learning models are designed in a two-step framework with offline encoders \cite{ilse2018abmil, shao2021transmil,lin2023ibmil, xiang2023ilramil, tang2024rrtmil, zhang2022dtfdmil}. Their feature representations lack task-specific knowledge, degrading pathological classification performance and interpretability. Among them, attention-based MIL frameworks introduce the inter-instance relationships. For example, ABMIL \cite{ilse2018abmil} selected the
top-k patches as positive instances by their attention scores. ACMIL \cite{zhang2024acmil} introduced multiple branch attention and stochastic top-k instance masking into attention challenging MIL to capture different instances with attention values. Besides, causality-based MIL models have been explored to capture instance correlations. IBMIL \cite{lin2023ibmil} addressed the spurious correlations between bags and labels based on the backdoor adjustment. CIMIL \cite{lin2024cimil} is a model-agnostic boosting MIL based on counterfactual inference, which employs instance searching strategy to help to localize reliable
instances. Another type of MIL models focus on feature refinement approaches. DGR-MIL \cite{zhu2024dgrmil} employed a diverse
global representation strategy to model the diversity among instances through a set of
global vectors. RRT-MIL \cite{tang2024rrtmil} proposed a re-embedded regional transformer to fine-tuning degraded features extracted from offline encoders. In general, these MIL models are all trained in a two-step framework, which lacks domain-specific knowledge and active interpretability.

\begin{figure*}[!t]
\centering
{\includegraphics[width=7in]{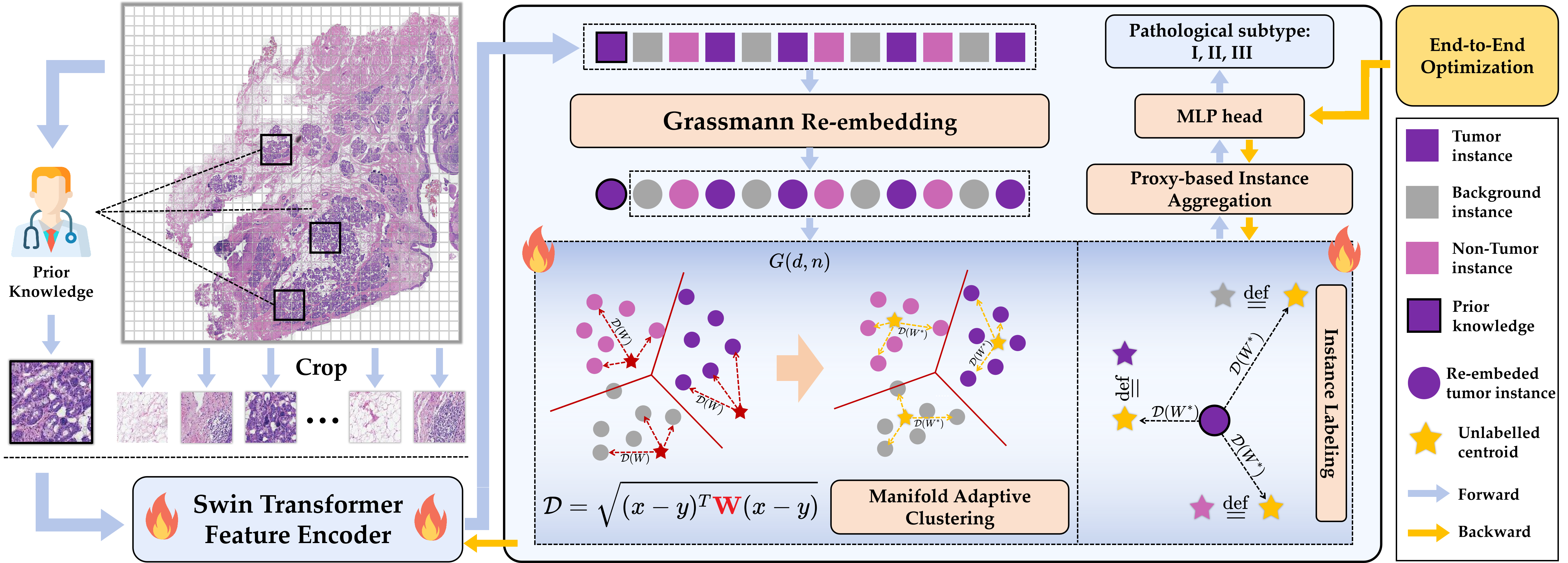}}
\hfil
\caption{Overall flowchart of proposed end-to-end MacNet. Prior knowledge instances are annotated by pathologists.}
\label{fig:fig2}
\end{figure*}

\subsection{Deep Clustering for Pathological Application}
Deep clustering methods have been widely explored for pathological application and beyond. As an unsupervised learning algorithm, they can offer more model interpretablity compared to deep learning models. Liu \cite{liu2025fuzzymil} \textit{et al.} proposed a deep fuzzy clustering framework with learnable parameters
that can decouple the morphological features of WSIs. Cluster-Siam \cite{wu2023improving} retains only the most relevant and interpretable clusters by implementing a cluster loss to guide effective representations. Contrastive learning is also employed for WSI analysis. For instance, Zhang \cite{zheng2024partial} \textit{et al.} designed a partial-label contrastive clustering (PLCC) module for label disambiguation to sample the most representative features of each category. Moreover, cDP-MIL \cite{chen2024cdp} models the instance-to-bag structure in WSIs using a cascade of Dirichlet processes that clusters instances based on feature covariance. CIMIL \cite{lin2024cimil} mentioned before also utilized K-means algorithm to obtain the results for counterfactual inference. These cluster-incorporated frameworks either use basic clustering method directly in high dimension, or have ambiguous clustering centroids that overemphasize low-effect and irrelevant background regions, thus degrading interpretability and classification performance.

\section{Methodology}

\subsection{Problem Statement}

\noindent \textbf{Basic Assumption:} Our proposed method is based on two assumptions. Firstly, the differentiation state in a pathological WSI is not uniform, so the pathological classification tasks meet the criteria of clustering-based MIL. Secondly, tumor instances are more correlated with pathological grading than non-tumor instances.

\noindent \textbf{Theoretical Analysis:} The goal of WSI-based classification task in MIL framework is to learn a permutation-invariant scoring function \cite{zaheer2017deep} $\mathcal{T}$ that maps the set of WSI instances $\mathcal{X}=\{x_1,\dots,x_N\}$ to label $\mathcal{Y}$. The decision-making process can be formulated as follows:
\begin{equation}
    \mathcal{Y} \leftarrow fc(\sigma(\mathcal{F}(\mathcal{X}))=\mathcal{T}(\mathcal{X})
\end{equation}
where $fc(\cdot)$ denotes the classifier layer, $\sigma(\cdot)$ denotes the aggregation fuction and $\mathcal{F}(\cdot)$ denotes the feature encoder.

As a quantitative measure of data uncertainty, information entropy can be used as an evaluation criterion for the theoretical performance of WSI-based pathological classification. For trainable feature encoder in our end-to-end framework, instance features become random variables. Let $\{\mathfrak{t}_1,\dots,\mathfrak{t}_N\}$ be the individual random variables in one WSI bag. Then the following theorem holds \cite{shao2021transmil}:

\textbf{Theorem 1}. \textit{The joint information entropy can be expressed as $\sum_{k=1}^{N}\mathcal{H}(\mathfrak{t}_k)$ iff. $\mathfrak{t}_k$ are i.d.d. random variables. Furthermore, when they are not i.d.d., we can derive that:
\begin{equation}
    \begin{aligned}
        \mathcal{H}(\mathfrak{t}_1,\cdots,\mathfrak{t}_N)&= \mathcal{H}(\mathfrak{t}_1) + \sum_{k=2}^{N}\mathcal{H}(\mathfrak{t}_k|\mathfrak{t}_1,\cdots,\mathfrak{t}_{k-1})\\
        &\le \sum_{k=1}^{N}\mathcal{H}(\mathfrak{t}_k)
    \end{aligned}
\end{equation}}

By the first assumption, the instances in the WSI bag have unique differentiation states that usually can be categorized as tumor instances (TIs), non-tumor instances (NTIs) and background instances (BGIs). Also, by introducing the prior knowledge instances (PKIs) from pathologists, we can model the correlation among different types of instances into information entropy as follows:

\begin{equation}\label{eq3}
    \setlength\abovedisplayskip{6pt}
    \setlength\belowdisplayskip{6pt}
    \begin{aligned}
        \mathcal{H}^*(\mathfrak{t}_1,\cdots,&\mathfrak{t}_N)= \sum_{k=1}^{n_1}\mathcal{H}(\mathfrak{t}_k | \{\mathfrak{t}^{\text{BGIs}}, \mathfrak{t}^{\text{NTIs}}, \mathfrak{t}^{\text{PKIs}}\}:\mathfrak{t}_k\in\mathfrak{t}^{\text{TIs}})\\
        &+\sum_{k=n_1+1}^{n_2}\mathcal{H}(\mathfrak{t}_k|\{ \mathfrak{t}^{\text{BGIs}}, \mathfrak{t}^{\text{TIs}}, \mathfrak{t}^{\text{PKIs}} \} :\mathfrak{t}_k\in\mathfrak{t}^{\text{NTIs}})\\
        &+\sum_{k=n_2+1}^{N}\mathcal{H}(\mathfrak{t}_k|\{ \mathfrak{t}^{\text{NTIs}}, \mathfrak{t}^{\text{TIs}}, \mathfrak{t}^{\text{PKIs}} \} :\mathfrak{t}_k\in\mathfrak{t}^{\text{BGIs}})\\
        &\le \mathcal{H}(\mathfrak{t}_1)+\sum_{k=2}^{N}\mathcal{H}(\mathfrak{t}_k|\mathfrak{t}_1,\cdots,\mathfrak{t}_{k-1}) \le \sum_{k=1}^{N}\mathcal{H}(\mathfrak{t}_k)
    \end{aligned}
\end{equation}
\noindent where $\mathcal{H}^*$ denote the correlated information entropy; $n_1$, $n_2-n_1$, $N-n_2$ denote the number of $\mathfrak{t}^{\text{TIs}}$, $\mathfrak{t}^{\text{NTIs}}$, $\mathfrak{t}^{\text{BGIs}}$ respectively. Conditioning cannot increase entropy $\mathcal{H}(X \mid Z)\le \mathcal{H}(X)$, where \(Z\) denotes the instance‐type labels from clustering. Thus, information entropy of the set of these random variables in \eqref{eq3} decreases after clustering stage. Theoretically, it indicates that the model can distinguish different instances robustly, thereby enhancing its ability to perform subtype classification tasks.

\subsection{Overview of Proposed MacNet}
The proposed MacNet aims to capture cross-instance correlations and the most influential instances related to pathological grading. Figure \ref{fig:fig2} illustrated the overall pipeline of our MacNet. Firstly, the cropped instances and prior knowledge instances are encoded into feature representation by learnable Swin Transformer. Secondly, we map these feature representations into a shared manifold subspace via grassmann re-embedding module. Then, we perform a novel manifold-constrained adaptive clustering strategy with a learnable matrix. Guiding by prior knowledge, we label the fuzzy clustered groups into different categories as TIs, NTIs and BGIs. Finally, we train the classifier using the ultimate feature representation from proxy-based instance aggregation. These modules combined can effectively enhance pathological classification performance and interpretability via one step end-to-end optimization.

\begin{figure}[!t]
\centering
\includegraphics[width=3.2in]{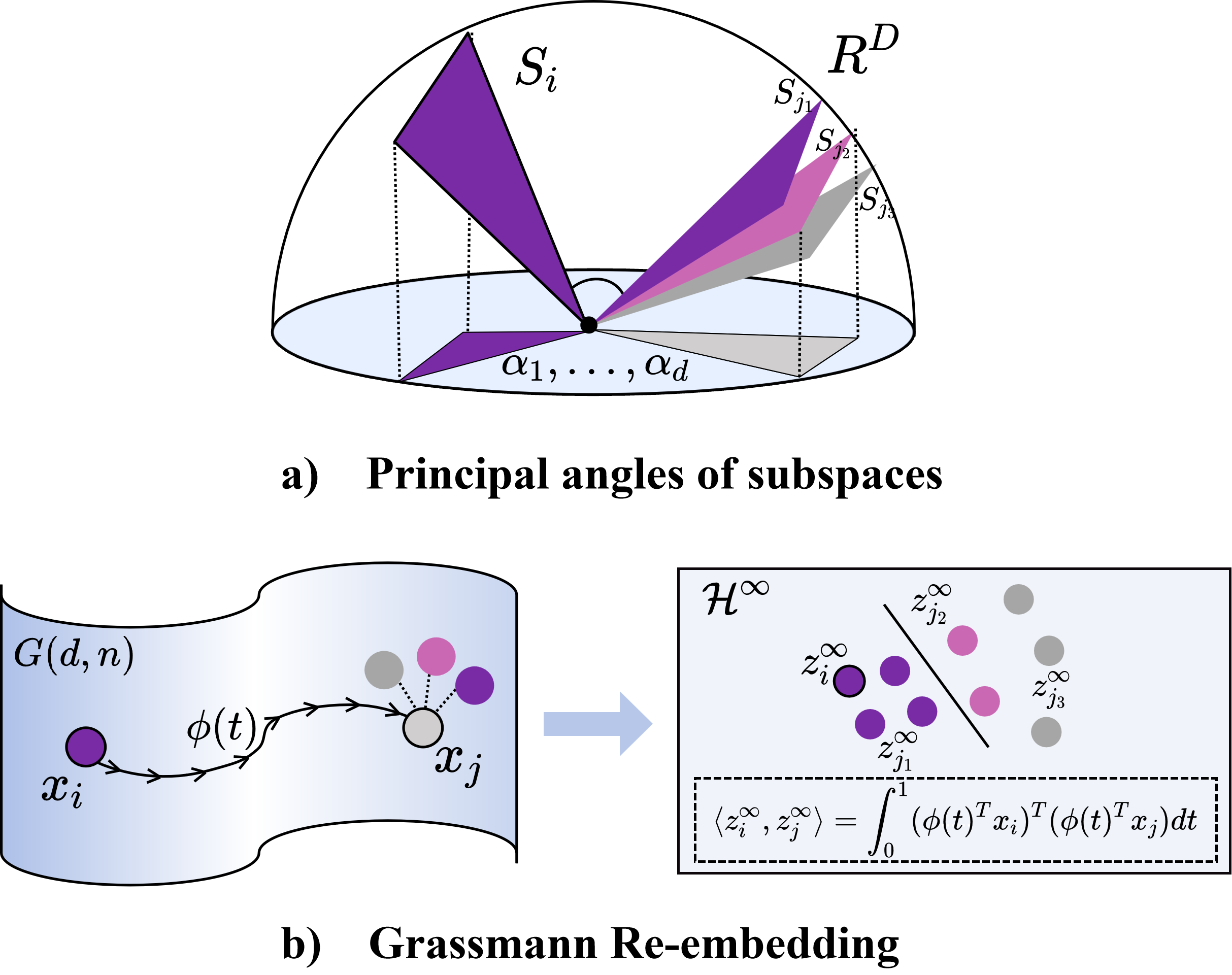}
\caption{\textbf{a):} Geodesic distance calculated by discrete principal angles. \textbf{b):} Grassmann Re-embedding maps features into infinite dimensional space. \label{fig:fig3}}
\end{figure}

\subsection{Grassmann Re-embedding}
Sampled WSI data lie on some low-dimensional
manifolds embedded in high-dimensional spaces. The features embedded in these subspaces share similar geometric properties, which naturally facilitates robust clustering. Thus, a well-defined feature transformation from original space into manifold space can contribute to classification performance.
 
\textbf{Definition 1}. (Grassmann Manifold) \textit{Grassmann manifold $G(d,n)$ is the space of all d-dimensional subspaces of a given n-dimensional vector space, i.e. each d-dimensional subspace $V$ is regarded as one basic element of $G(d,n)$}.

From the definition above, we notice that eigenvector spaces computed from principal component analysis (PCA) can be regarded as points in Grassmann manifold. One direct approach to measure the degree that subspaces overlap is using principal angles \cite{hamm2008grassmann}, where the distance metric $\|\boldsymbol{\alpha}\|_2$ uses only two discrete subspace. It is not aligned with the manifold assumption with locally Euclidean property; see Figure \ref{fig:fig3} a). 

Inspired by the discussion above and the idea of domain adaptation in GFK \cite{wang2018meda}, we first utilize PCA algorithm to compute the subspaces $S_i$, $S_j$ for prior knowledge source domain of PKIs, unlabelled target domain of $\{\text{TIs},\text{NTIs},\text{BGIs}\}$, respectively. Specially, we denote column-orthogonal matrices $\mathbf{P}_i,\mathbf{P}_j\in\mathbb{R}^{n\times d}$ as two basis of $S_i$, $S_j$. Therefore, the geodesic flow $\{\phi(t):0\le t\le1\}$ between two points in $G(d,n)$ can form a path from $\mathbf{P}_i=\phi(0)$ to $\mathbf{P}_j=\phi(1)$. It also approximates the actual geodesic distance in Grassmann manifold by mapping original features into space with infinite dimension; see Figure \ref{fig:fig3} b) and \eqref{eq5}.
\begin{equation}\label{eq5}
    \langle z_i^{\infty}, z_j^{\infty} \rangle=\int_0^1(\phi(t)^Tx_i)^T(\phi(t)^Tx_j)dt=x^T_i\mathbf{G}x_j
\end{equation}
where kernel $\mathbf{G}\in \mathbb{R}^{n\times n}$ is a positive semi-definite matrix. If $X_i\in \mathbb{R}^{N'\times n},X_j\in \mathbb{R}^{N\times n}$ denote PKIs and unlabelled WSI bag feature matrices, re-embedding feature $Z^{\infty}_i,Z^{\infty}_j$ in the infinite-dimensional space are calculated by $Z^{\infty}=X\sqrt{\mathbf{G}}$.

Note that the original expression of kernel $\mathbf{G}\in \mathbb{R}^{n\times n}$ is derived by generalized singular value decomposition \cite{gong2012geodesic}. To incorporate Grassmann re-embedding module into our framework, we innovated the underlying derivation and implementation for end-to-end optimization. See Appendix.

\begin{algorithm}[!t]
\small
\caption{Proxy Instance Labeling and Aggregation}
\label{alg:psla}
\begin{algorithmic}[1]
\Require Feature clusters \(\mathbf{Z}=\{Z_{j_1}^{\infty},Z_{j_2}^{\infty},Z_{j_3}^{\infty}\}\), Prior Knowledge \(Z_i^{\infty}\)
\Ensure  Final bag feature \(Z_{final}\) and label map \(\texttt{label}\)
\For{\(k=1\) to \(3\)}   
  \State \(M_k=\mathtt{mean}(Z_{j_k}^{\infty},\mathtt{dim=0})\)
  \State \(M_p=\mathtt{mean}(Z_i^{\infty},\mathtt{dim=0})\)
  \State \(d_k \;\gets\; \mathcal{D}\bigl(M_k,M_p)\) \Comment{Adaptive distance metric}
\EndFor
\State \(\{w_k\}_{k=1}^3 \gets 1-\mathtt{Normalize}(\{d_k\})\)
\State \(\texttt{order}\gets\mathtt{argsort}(\{w_k\},\,\texttt{descend})\)
\State \(\texttt{label}[\texttt{TIs}]\gets \mathbf{Z}_{\texttt{order}[1]},\;\dots,\;\)
  \(\texttt{label}[\texttt{BGIs}]\gets \mathbf{Z}_{\texttt{order}[3]}\)
\State \(Z_{ final}\gets \texttt{concat}\Big( (\sum_{k=1}^3 w_k\ Z_{j_k}^{\infty}),Z_{i}^{\infty}\Big)\) 
\State \Return \(Z_{final},\,\texttt{label}\)
\end{algorithmic}
\end{algorithm}

\subsection{Manifold Adaptive Clustering}
According to the concept above, features re-embedded into Grassmann manifold can contribute to clustering process due to the benign geometric properties. However, main-stream clustering methods with $L_p$ norm are designed to assign equal weight to each feature dimension. Inspired by the idea of metric learning, we proposed an adaptive clustering approach under manifold constraint to solve the problems above.

The general form of metric learning is based on Mahalonabis distance in the following expression.
\begin{equation}\label{eq6}
    \begin{aligned}
        \mathcal{D}(x,y)=\sqrt{(x-y)^T\mathbf{W}(x-y)}\\
    \mathbf{W}=diag(w_1,w_2,\dots,w_n),\quad w_k\geq0
    \end{aligned}
\end{equation}
where $\mathbf{W}\in \mathbb{R}^{n\times n}$ is a positive semi-definite matrix and $x,y \in \mathbb{R}^{n\times 1}$ denote the instance feature representation vectors. If $\mathbf{W}$ is an identity matrix, the distance degrades as standard Euclidean distance.

The Grassmann re-embedding process performs a feature mapping by calculating the geodesic distance between column-orthogonal subspaces. This reduces the feature interaction within different dimensions and focuses more on the effect of features along each individual direction. Therefore, we only need to keep the matrix $\mathbf{W}$ diagonal with trainable parameters in \eqref{eq6}. Subsequently, we replace the $L_p$ norm with the adaptive distance $\mathcal{D}(x,y)$ in the conventional KMeans clustering algorithm; see Figure \ref{fig:fig2}. Through end-to-end optimization, the ultimate $\mathbf{W}$ can assign different weights to each dimension and capture the most important ones, thus achieving robust and adaptive clustering.

\subsection{Proxy Instance Labeling and Aggregation}
After manifold adaptive clustering process, the labels of clustering centroids still remain ambiguous. Hence, it will inhibit the knowledge of TIs and overemphasize low-effect NTIs and irrelevant BGIs. Guiding by pathologists' domain knowledge, we proposed a proxy instance labeling and aggregation strategy; see Figure \ref{fig:fig2} and Algorithm \ref{alg:psla}. 

After the clustering module, we categorize the input WSI features into three groups as $Z_j^{\infty}=\{Z^{\infty}_{j_1},Z^\infty_{j_2},Z^\infty_{j_3}\}$. Let $M_1,M_2,M_3$ denote their mean centroids, respectively. 
\begin{equation}\label{eq7}
    \begin{aligned}
        M_k &= \text{mean}(Z^{\infty}_{j_k},\text{dim}=0),\quad k=1,2,3\\
        M_p&=\text{mean}(Z^{\infty}_{i},\text{dim}=0)
    \end{aligned}
\end{equation}
where $\text{mean}(\cdot,\text{dim})$ returns the mean vector of feature along certain dimension. $M_k$ is regarded as proxy instance in each group and we obtain the proxy instance of PKIs in \eqref{eq7} too.

Followed by the trainable weight $\mathbf{W}$ and metric $\mathcal{D}(\cdot,\cdot)$ in manifold adaptive clustering process, we can further estimate the pseudo labels by the following equations.
\begin{align}\label{eq8}
    \begin{cases}
        d_{k}=\mathcal{D}(M_k,M_p)\\
        Z_{\text{tis}}^{\infty} \triangleq Z_{j_{\text{min}}}^{\infty},\quad \text{s.t.}\quad \text{min} ={{\operatorname{argmin}_k\{d_k\}}}   \\
        Z_{\text{bgis}}^{\infty} \triangleq Z_{j_{\text{max}}}^{\infty},\quad \text{s.t.}\quad \text{max} ={{\operatorname{argmax}_k\{d_k\}}}   \\
       Z_{\text{ntis}}^{\infty}  \triangleq Z_j^{\infty}\setminus \{Z_{\text{tis}}^{\infty},Z_{\text{bgis}}^{\infty}\}
    \end{cases}
\end{align}
where the shortest distance labels the most relevant tumor instances and the longest distance labels background.
\begin{equation}\label{eq9}
    Z_{final}=(Z^{\infty}_{\text{tis}},Z^\infty_{\text{bgis}},Z^\infty_{\text{ntis}},Z^\infty_{i})\cdot 
\begin{pmatrix}
1-d_{\text{min}} \\
1-d_{\text{max}} \\
1-d_{\text{med}} \\
1
\end{pmatrix}
\end{equation}
where $d_{\text{med}}$ denotes the median value of $\{d_1,d_2,d_3\}$.

Finally, we assign learnable weights in \eqref{eq9} and minimizing the loss function $\mathcal{L}=\mathcal{L}_{ce}(\sigma(Z_{final}))+d_{\text{min}}-d_{\text{max}}$ ($\sigma(\cdot)$ denote aggregation function with classifier head and the last two regularization terms facilitate clustering). Since KMeans clustering is not normally differentiable, these two terms aim to optimize the trainable matrix $\mathbf{W}$ by computing the gradients $\frac{\partial \mathcal{L}}{\partial \mathbf{W}}$. In this case, we introduce the quantitative correlation within three instance groups, which adaptively filter out unimportant
information through end-to-end optimization. Therefore, it leads the contributive information to improve the performance of classification and interpretability.

\section{Experiments and Results}
\subsection{Datasets and Setup}
\noindent \textbf{Datasets:} To evaluate proposed MacNet for diagnosis and sub-typing tasks, we conduct extensive experiments on two public datasets and two private datasets. CAMELYON16 \cite{bejnordi2017camelyon16} is for breast cancer lymph node metastasis detection. DHMC-LUNG \cite{wei2019lung} is for lung adenocarcinoma classification. Two private WSI datasets were collected from Army Medical University, where AMU-LSCC is for laryngeal squamous cell carcinoma pathological grading and AMU-CSCC is for cervical squamous cell carcinom. AMU-LSCC dataset includes 342 whole slide images that was divided by a 6:4 training-validation ratio for each category. Specially Grade I contains a total of 89 WSIs, Grade II contains 152 WSIs and Grade III includes 101 WSIs. AMU-CSCC dataset includes 262 whole slide images, where Grade I contains a total of 27 WSIs, Grade II contains 127 WSIs and Grade III includes 108 WSIs. 961 patches were grid cropped from each WSI on two datasets. LSCC includes 81 prior knowledge instances per bag and CSCC includes 64 prior knowledge instances per bag. For CAMELYON16 and DHMC-LUNG, 5
tumor-guiding instances are added into each bag.

\noindent \textbf{Compared Models:} We employ eleven different types of SOTA models for comparison, which are ABMIL \cite{ilse2018abmil}, CLAMs \cite{lu2021clam}, IBMIL \cite{lin2023ibmil}, DGRMIL \cite{zhu2024dgrmil}, DTFD-MIL \cite{zhang2022dtfdmil}, ILRA-MIL \cite{xiang2023ilramil}, TransMIL \cite{shao2021transmil}, ACMIL \cite{zhang2024acmil}, S4MIL \cite{fillioux2023s4mil}, RRT-MIL \cite{tang2024rrtmil} and MFC-MIL \cite{mfcmil2025}. All compared models are designed in a two-step framework with offline feature encoders. To comprehensively compare their classification capability, we extract features with Swin Transformer \cite{liu2021swin} pretrained on ImageNet-11k and PLIP \cite{huang2023plip} pretrained on OpenPath, respectively. Moreover, we report Accuracy and Area Under Curve (AUC) for evaluation metrics.

\noindent \textbf{Implementation Details:} To ensure reproducibility, all compared models are employed under the same experimental configurations using PyTorch on four \textit{NVIDIA A10 Tensor Core 24GB} We utilize Rmsprop optimizer for end-to-end learning. Models on two private datasets were trained with a batch size of 2 and 100 epochs. For two public datasets, batch size is set 1. The random seed was fixed 0 across all stages. The input resolution of MacNet was $96\times 96 \times3$. The learning rate schedule was: $1\times10^{-5}$ for epochs 1–50, $5 \times 10^{-6}$ for epochs 51–75, and $1 \times 10^{-6}$ for epochs 76–100.

\subsection{Comparison with SOTA Methods}
For diagnosis tasks like CAMELYON16 (tumor or normal) , there still exists various differentiation states in a slide, where it can be more accurately diagnosed when patches are clearly categorized. Table \ref{tab:table1} illustrates the pathological diagnosis and sub-typing performance of MIL approaches on two private datasets and two public datasets. The results suggest that our proposed MacNet outperforms all SOTA models under all three evaluation metrics and all four benchmarks. For features pretrained on ImageNet-11k, ACMIL with multi-head attention achieves the best classification performance on AMU-LSCC except MacNet, which improves the ACC and AUC by 5.07\% and 4.38\%, respectively. On CAMELYON16 dataset, the improvement is 8.57\% and 10.2\% compared to DGR-MIL. Other two datasets have seen the similar trend with substantial enhancement of 0.90\% AUC improvement on AMU-CSCC and 5.49\% AUC improvement on DHMC-LUNG.

\begin{figure*}[!t]
\centering
{\includegraphics[width=6.92in]{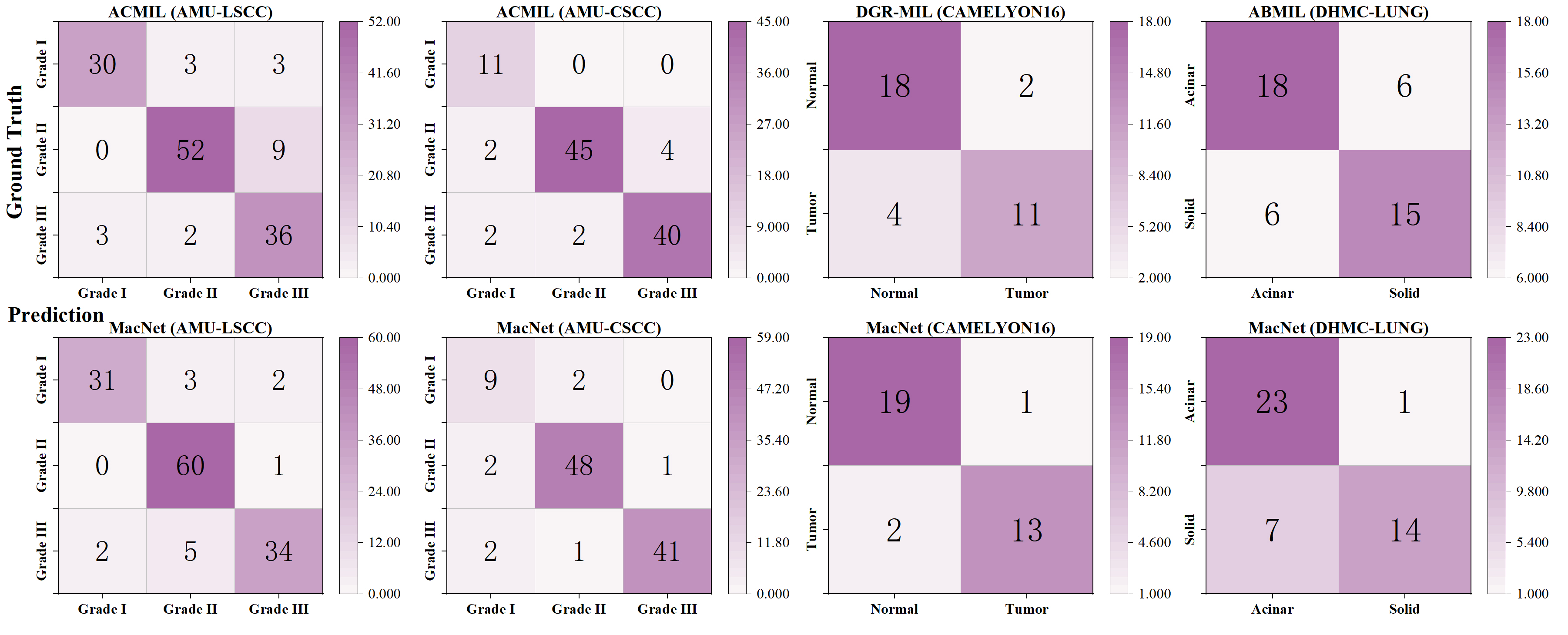}}
\hfil
\caption{Confusion Matrix plots on multicentre datasets.}
\label{fig:confux}
\end{figure*}

It is worth noting that although incorporating features extracted by PLIP, these models fail to exhibit a generally significant improvement. Only performance on CAMELYON16 shows a comprehensive gain with the second-best ACC from 82.86\% to 85.71\% and AUC from 82.53\% to 85.88\%. Despite this, our MacNet is superior to all compared SOTA models. Moreover, the confusion matrices in Figure \ref{fig:confux} show that our MacNet can distinguish more categories than the sub-optimal SOTA models. Also, the ROC plots from Figure \ref{fig:roc} illustrate that the curve for MacNet lie above all other curves, implying the largest AUC value and strongest classification performances. We attribute this achievement to the advantages of clustering-based MIL framework with manifold constraints and end-to-end optimization, which will be validated in the next section.

\begin{table*}[!t]
\small
\centering
\renewcommand\arraystretch{1.1}
\caption{Performance comparison of cancer diagnosis, and sub-typing results on two private and two public datasets.\label{tab:table1}}
\begin{tabular}{clcccccccc}
\Xhline{1pt}
& \multirow{2}*{\normalsize Models} & \multicolumn{2}{c}{AMU-LSCC} & \multicolumn{2}{c}{AMU-CSCC} & \multicolumn{2}{c}{CAMELYON16} & \multicolumn{2}{c}{DHMC-LUNG}\\ \cmidrule(r){3-4} \cmidrule(r){5-6} \cmidrule(r){7-8} \cmidrule(r){9-10}
& ~ & \makebox[0.06\textwidth][c]{ACC}  & \makebox[0.06\textwidth][c]{AUC} & \makebox[0.06\textwidth][c]{ACC} & \makebox[0.06\textwidth][c]{AUC} & \makebox[0.06\textwidth][c]{ACC}  & \makebox[0.06\textwidth][c]{AUC} & \makebox[0.06\textwidth][c]{ACC}  & \makebox[0.06\textwidth][c]{AUC} \\ \hline
{\multirow{12}*{\rotatebox{90}{\textbf{Swin-T on ImageNet-11k}}}}& ABMIL \textit{(ICML'18)}& 0.6739 & 0.7888 & 0.8302  & 0.9547 &0.7429&0.7918&0.7333&\underline{0.8123} \\
~ & CLAM-SB\textit{(N.B.E.'21)}& 0.6957  & 0.8367 & 0.8302  & 0.9581 &0.7429&0.7478&0.7556&0.7783 \\
~& CLAM-MB \textit{(N.B.E.'21)}& 0.6812  & 0.8031 & 0.8774 & 0.9336 &0.7429&0.7339&0.7333&0.7684 \\
~& TransMIL \textit{(ICCV'21)}& 0.8261  & 0.9021 & 0.8868 & 0.9521&0.7429&0.6898&0.7333&0.8084 \\
~& DTFD-MIL \textit{(CVPR'22)}& 0.5435  & 0.7245 & 0.8113  & 0.9036 &0.7143&0.7306&0.7333&0.8084 \\
~& IBMIL-DS \textit{(CVPR'23)}& 0.6377  & 0.7672 & 0.7830  & 0.8881 &0.7429&0.7037&0.6889&0.7091 \\
~& ILRA-MIL \textit{(ICLR'23)} & 0.6884  & 0.8048 & 0.8491  & 0.9123 &0.7714&0.7306&0.7778&0.8074  \\
~& S4MIL \textit{(MICCAI'23)}& 0.7971 & 0.8845 & 0.8491  & 0.9605  &0.7429&0.7322&0.7111&0.7007\\
~& DGR-MIL \textit{(ECCV'24)}& 0.8188  & 0.9147 & 0.8868  & 0.9613 &\underline{0.8286}&\underline{0.8253}&\underline{0.8000}&0.7886  \\
~& ACMIL-MHA \textit{(ECCV'24)}& \underline{0.8551}  & \underline{0.9219} & \underline{0.9057}  & 0.9701  &0.8000&0.7331&0.6889&0.6849\\ 
~& RRT-MIL \textit{(CVPR'24)} & 0.7681 & 0.8732 & 0.8585 & 0.9265  &0.7143&0.7608&0.7778&0.7521\\
~& MFC-MIL \textit{(ICLR'25)} & 0.8261 & 0.9126 & 0.8962 & \underline{0.9723}  &0.7714&0.7208&0.7778&0.7767\\ \hline
~& \textbf{MacNet (ours)} & \textbf{0.9058}  & \textbf{0.9657} & \textbf{0.9245}  & \textbf{0.9813} &\textbf{0.9143}&\textbf{0.9273}&\textbf{0.8222}&\textbf{0.8672}\\
\Xhline{1pt}
{\multirow{12}*{\rotatebox{90}{\textbf{PLIP on OpenPath}}}}& ABMIL\textit{(ICML'18)}& 0.7536  & 0.8689 & 0.8208  & 0.9321 &0.7429&0.7820&0.7333&0.7467 \\
~ & CLAM-SB \textit{(N.B.E.'21)} & 0.7246  & 0.8367 & 0.8774  & 0.9603 &0.7714&0.8392&\underline{0.7556}&0.7674 \\
~& CLAM-MB \textit{(N.B.E.'21)} & 0.7681  & 0.8556 & 0.8491  & 0.9333 &0.7714&0.8310&0.7333&0.7659 \\
~& TransMIL \textit{(ICCV'21)} & 0.8188  & 0.9159 & 0.8774 &  0.9640 &0.8000&0.8506&0.7333&0.7442 \\
~& DTFD-MIL \textit{(CVPR'22)} & 0.6739  & 0.8138 & 0.7736  & 0.8960 &0.6571&0.6514&0.6222&0.6267 \\
~& IBMIL-DS \textit{(CVPR'23)}& 0.7609  & 0.8738 & 0.8113  & 0.9080 &0.7714&0.7106&0.6889&0.7131 \\
~& ILRA-MIL \textit{(ICLR'23)} & \underline{0.8551}  & \underline{0.9199} & \underline{0.9057}  & 0.9713 &0.7714&0.7918&0.7333&0.7773  \\
~& S4MIL\textit{(MICCAI'23)}& 0.8188 & 0.9065 & 0.8585  & 0.9652  &0.7714&0.7780&0.6667&0.6983\\
~& DGR-MIL \textit{(ECCV'24)}& 0.8188  & 0.9060 & 0.8774  & 0.9569 &\underline{0.8571}&\underline{0.8588}&0.7333&\underline{0.7886}  \\
~& ACMIL-MHA \textit{(ECCV'24)}& 0.8406  & 0.9131 & 0.8868  & \underline{0.9735}  &0.8000&0.7094&0.7333&0.7635\\
~& RRT-MIL \textit{(CVPR'24)}& 0.8333  & 0.9057 & 0.8679  & 0.9543  &0.8286&0.8343&0.7111&0.7294\\ 
~& MFC-MIL \textit{(ICLR'25)} & 0.8261 & 0.9067 & 0.8868 & 0.9526  &0.8000&0.8367&0.7556&0.7575\\ \hline
~& \textbf{MacNet(ours)} & \textbf{0.9058}  & \textbf{0.9657} & \textbf{0.9245}  & \textbf{0.9813} &\textbf{0.9143}&\textbf{0.9273}&\textbf{0.8222}&\textbf{0.8672}\\
\Xhline{1pt}
\end{tabular}
\end{table*}

\begin{figure*}[!t]
\centering
{\includegraphics[width=6.5in]{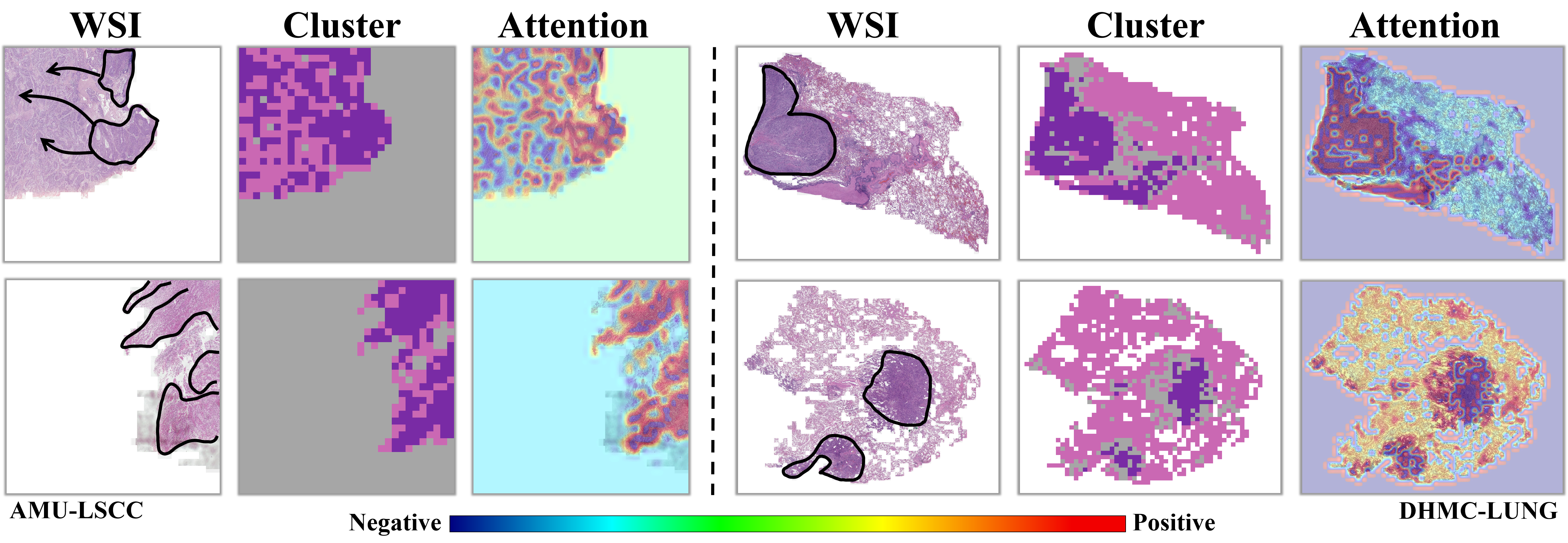}}
\hfil
\caption{Visualization results on AMU-LSCC and DHMC-LUNG. The first column is original WSI. The second column is labeling result during prediction. The last column is the attention heatmap corresponding to its category.}
\label{fig:fig5}
\end{figure*}

\begin{table}[!t]
\small
\centering
\renewcommand\arraystretch{1.1}
\caption{Ablation experiments on different modules of MacNet. The baseline model is vanilla Swin Transformer.\label{tab:table2}}
\begin{tabular}{lcccc}
\Xhline{1pt}
\multirow{2}*{Models}& \multicolumn{2}{c}{AMU-LSCC} & \multicolumn{2}{c}{CAMELYON16} \\ \cmidrule(r){2-3} \cmidrule(r){4-5} 
~ & ACC & AUC & ACC & AUC\\ \hline
ACMIL-MHA & 0.8551 & 0.9219 & -- & --\\
 DGR-MIL & -- & -- & 0.8286 & 0.8253\\\hline
 baseline & 0.8696 & 0.9567 & 0.8286 & 0.8286\\
+ L1-KMeans & 0.8986 & 0.9648 & 0.8857 & 0.8727 \\
+ L2-KMeans & 0.8913 & 0.9598 & 0.8286 & 0.8808 \\ \hline
\makecell[l]{+ PCA \\ \& Adaptive-KMeans}& 0.8841 & 0.9655 & 0.8857 & 0.8988 \\
\textbf{\makecell[l]{+ Grassmann \\ \& Adaptive-KMeans}} & \textbf{0.9058} & \textbf{0.9657} & \textbf{0.9143} & \textbf{0.9273}\\
\Xhline{1pt}
\end{tabular}
\end{table}

\begin{figure}[!t]
\centering
{\includegraphics[width=3.5in]{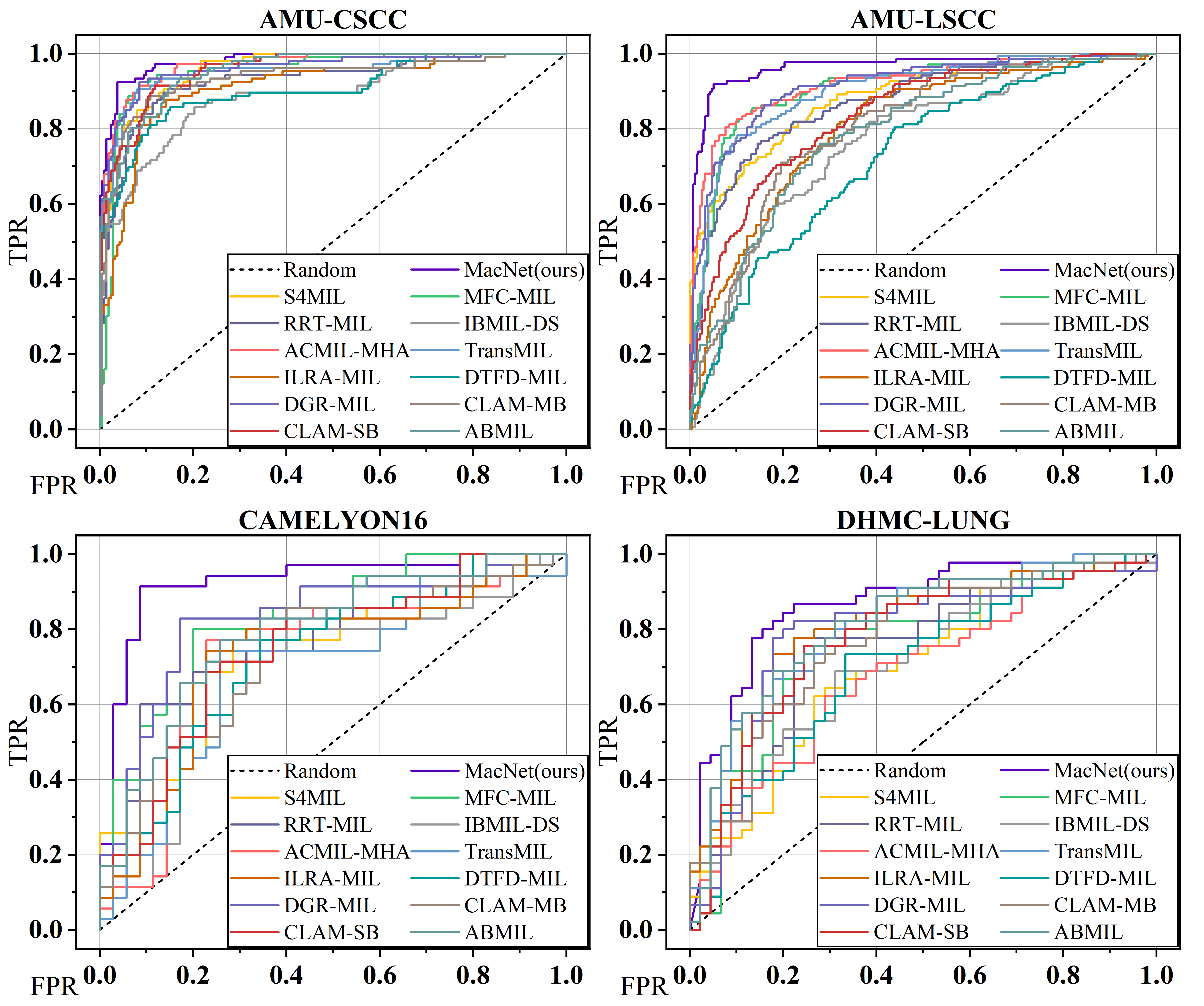}}
\hfil
\caption{ROC plots on multicentre datasets.}
\label{fig:roc}
\end{figure}

\vspace{-1em}
\subsection{Ablation Study}
\vspace{-0.3em}
Note that we fix cluster number to be 3 because coarse labels as tumor, non-tumor and background are actually enough for most WSI diagnosis tasks. Larger cluster centroids are needed for more complex tasks \cite{song2024fasial}. Subsequently,
we conducted ablation experiments to verify the effectiveness of modules proposed in MacNet; see Table \ref{tab:table2}. The baseline is a fully supervised mean-pooling framework with end-to-end optimization. Without injecting any domain knowledge or data-specific tricks, our baseline outperforms best ACMIL on AMU-LSCC and best DGR-MIL on CAMELYON16. It suggests that end-to-end learning can learn better task-related feature representation.

When incorporating clustering methods with prior knowledge, the models show a significant improvement. Specifically, the ACC and AUC metrics for L1-based KMeans increase by 5.71\% and 4.41\% on CAMELYON16, respectively. While slightly less effective than L1 distance, the L2 distance also exhibits considerable effectiveness, with 2.17\% ACC and 0.31\% AUC improvement on AMU-LSCC.

Compared to linear projection methods like PCA, Grassmann re-embedding can capture the nonlinear geometric structures in complex patterns, which plays a crucial role in clustering and classification tasks. Although PCA projection with adaptive KMeans achieves second-best AUC on both datasets. When coupling the two interacting modules, our proposed MacNet with manifold-constrained adaptive clustering achieves the best performance.

\begin{figure*}[!t]
\centering
{\includegraphics[width=6in]{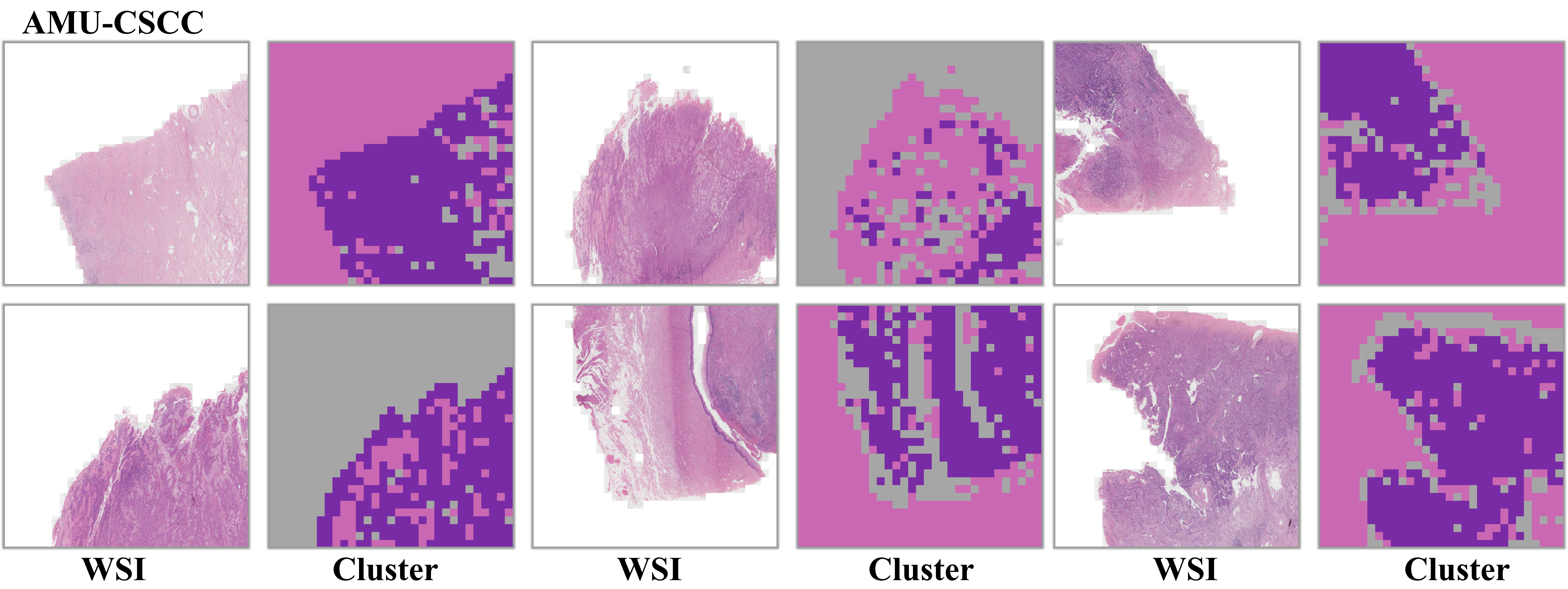}}
\hfil
\caption{Visualization results for AMU-CSCC from manifold adaptive clustering performing potential semantic segmentation.}
\label{fig:clu_cscc}
\end{figure*}

\begin{figure*}[!t]
\centering
{\includegraphics[width=6.25in]{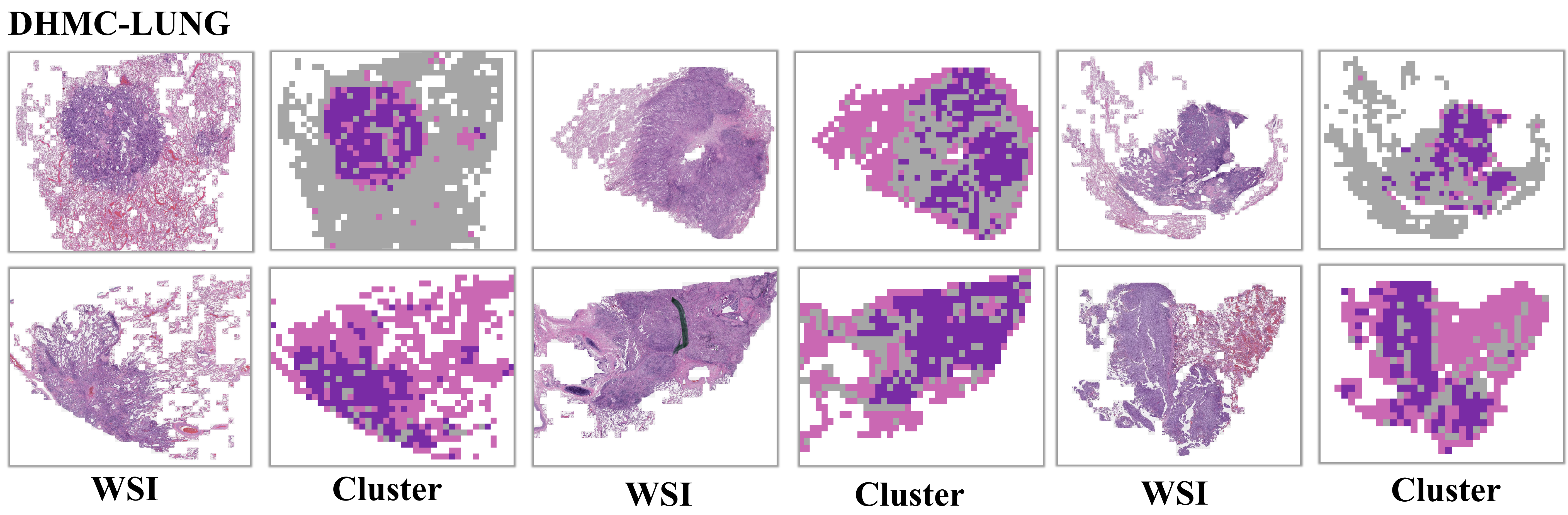}}
\hfil
\caption{Visualization results for DHMC-LUNG from manifold adaptive clustering performing potential semantic segmentation.}
\label{fig:clu_lung}
\end{figure*}

\begin{figure*}[!t]
\centering
{\includegraphics[width=7in]{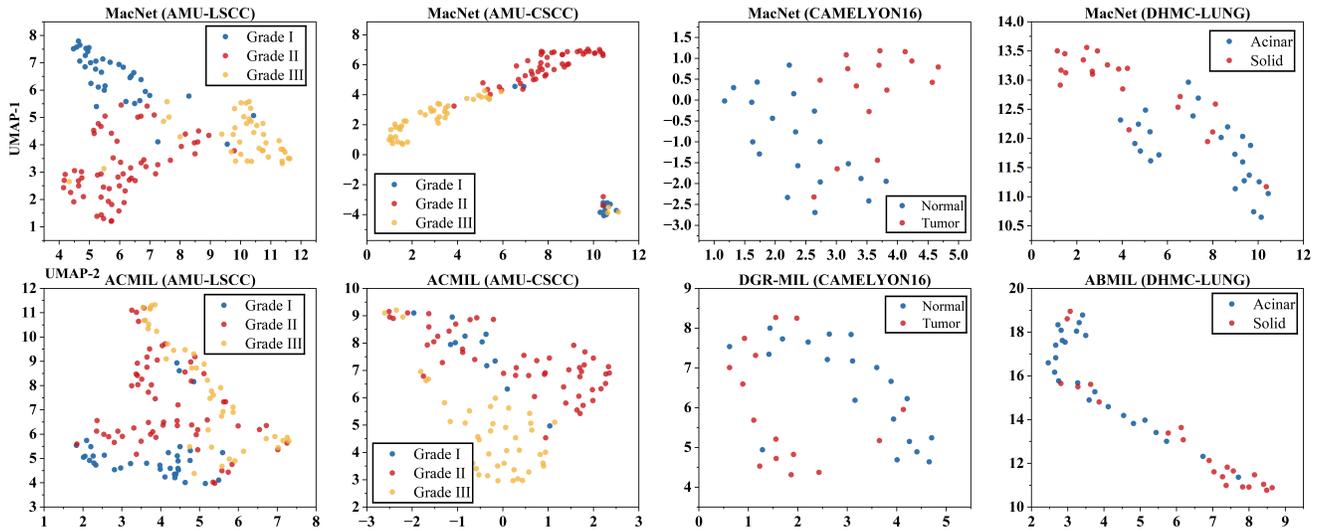}}
\hfil
\caption{2-dimensional feature representation by UMAP on multicentre datasets.}
\label{fig:2d}
\end{figure*}

\begin{figure}[!t]
\centering
{\includegraphics[width=3.5in]{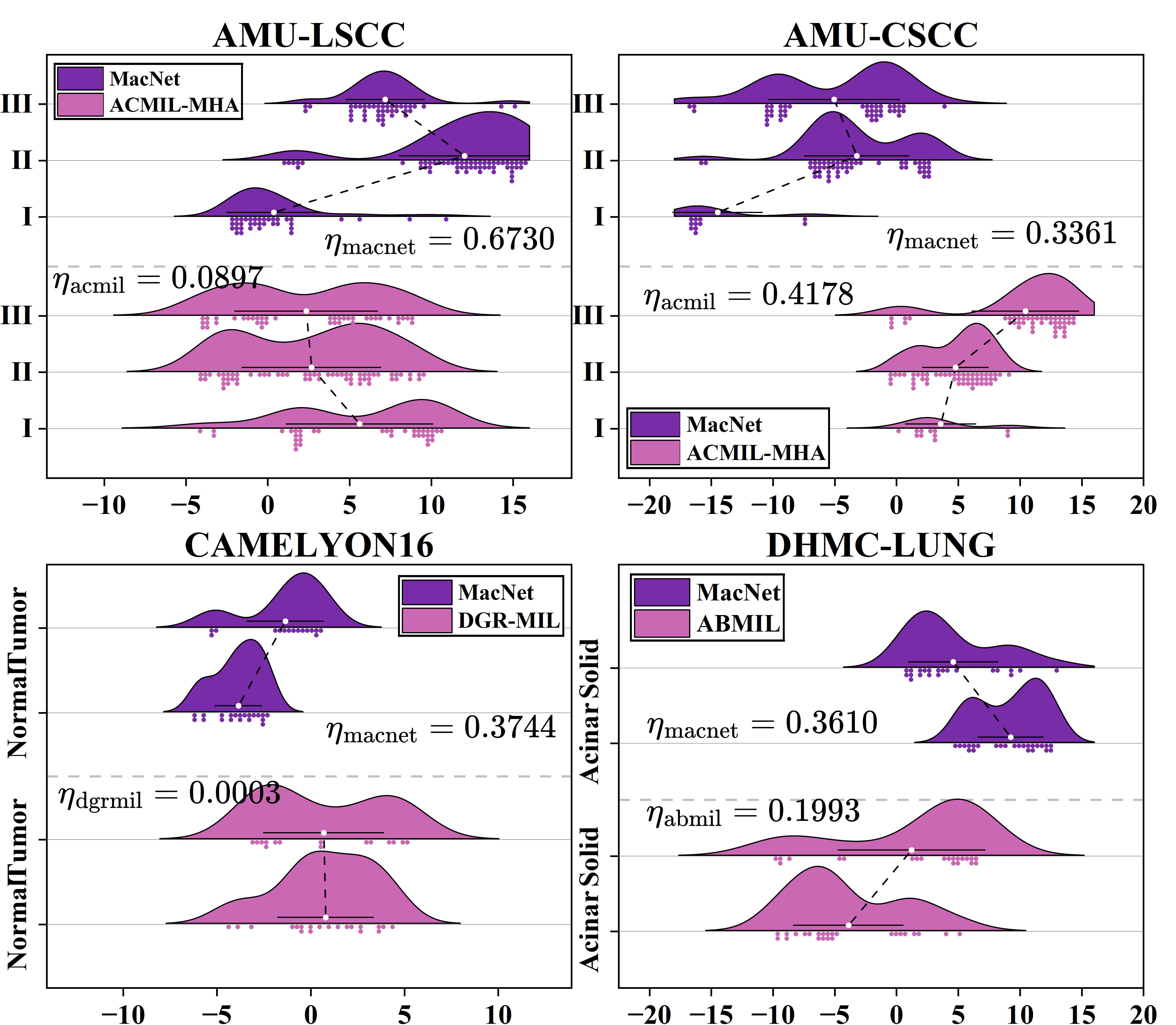}}
\hfil
\caption{ANOVA plots on multicentre datasets.}
\label{fig:anova}
\end{figure}

\subsection{Visualization of Model Interpretability}
As discussed in Introduction, clustering-incorporated framework provides better pathological motivation. Figure \ref{fig:fig5} illustrates the visualization results of MacNet in terms of clustering process and attention heatmap on AMU-LSCC and DMHC-LUNG. Both two results demonstrate a strong alignment with pathologists' annotation. The clustering process validate the model's active interpretability that approximates higher differentiation areas during decision-making. Showing the most contributive regions to the prediction, the attention heatmap suggests that our model has excellent passive interpretability. Overall, our MacNet exhibits a comprehensive and high level of interpretability. It also implies a potential paradigm for weakly supervised semantic segmentation by the clustering results in Figure \ref{fig:clu_cscc} and Figure \ref{fig:clu_lung}.

\subsection{Effects of Feature Representation Capability}

An effective classification model is typically capable of capturing and fitting the differences among various pathological patterns. To validate this hypothesis, we extract the feature representations from the final layer of the model. We then applied UMAP \cite{becht2019umap} for dimensionality reduction and used ANOVA visualizations with effect size metrics ($\eta^2$) to display the 1-dimensional feature embeddings.

The effect size in ANOVA indicates quantifies the proportion of variance in the dependent variable explained by the grouping factor. A larger effect size suggests a stronger ability of the model to capture and distinguish differences between classes. Figure \ref{fig:anova} shows that MacNet models the distinct distributions within different categories, while ACMIL and DGR-MIL tend to overlap. It has the larger effect size 0.6730, 0.3744 compared to the second-best models on AMU-LSCC and CAMELYON16, respectively. Same as other two datasets, our model has generally larger effect size on DHMC-LUNG, which exhibits
refined feature representation towards pathological classification. Although the effect size on AMU-CSCC is smaller
than the second-best model, it still has a generally excellent
above average level. 

In addition, we visualize the feature representation in 2D dimension by UMAP. In Figure \ref{fig:2d}, MacNet represents relatively clear decision boundaries on multicentre datasets while other sub-optimal SOTA models have many overlapping samples, thereby degrading the classification performance.

Therefore, our proposed MacNet exhibits better feature feature representation ability.

\begin{figure}[!t]
\centering
{\includegraphics[width=3.5in]{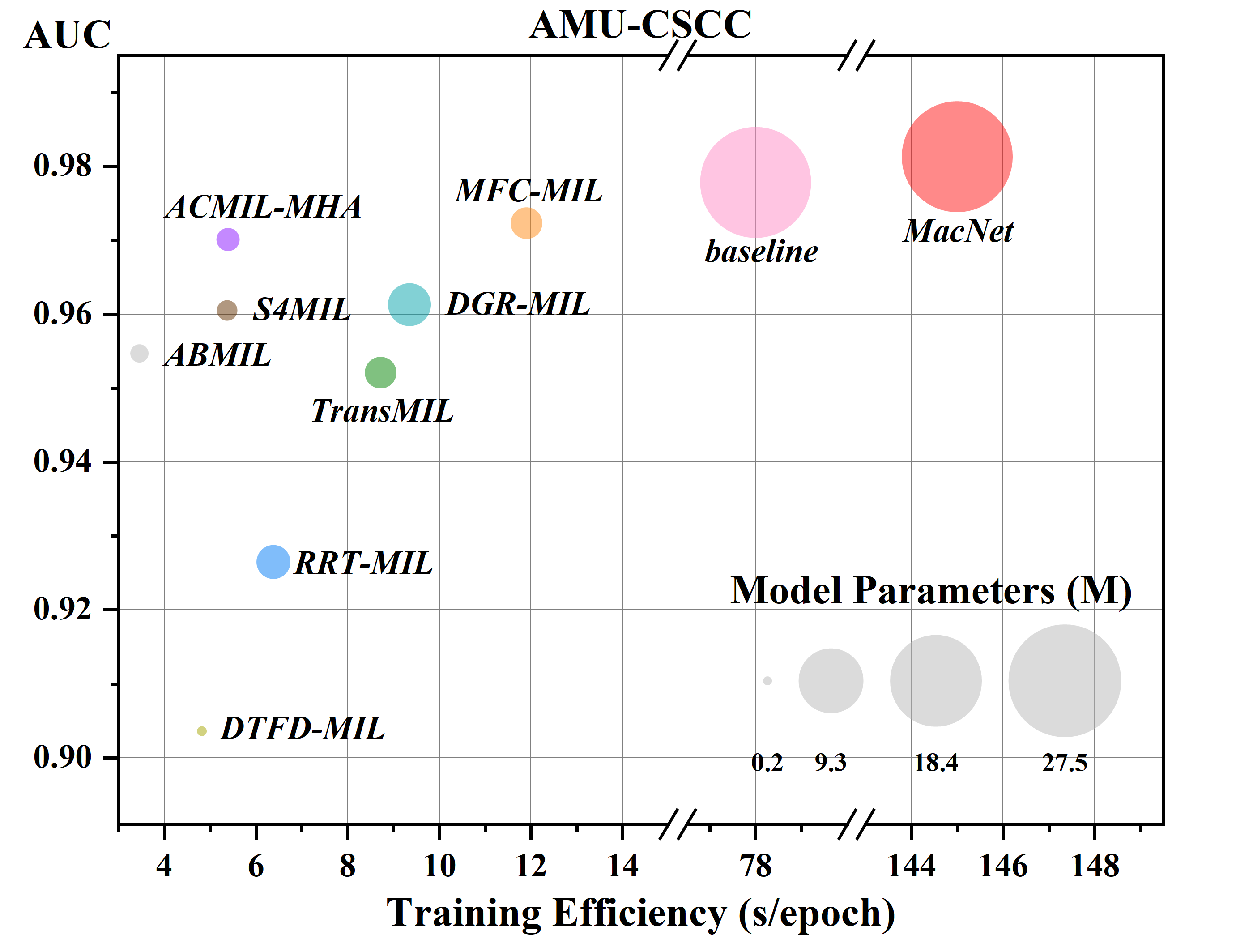}}
\hfil
\caption{Comparison on model complexity and training efficiency (per epoch) on AMU-CSCC.}
\label{fig:complex}
\end{figure}

\subsection{Model Complexity and Training Efficiency}
 End-to-end learning for WSI analysis is generally considered computationally demanding, thus receiving limited attention. However, we found that the computational cost has been significantly overestimated. In Figure \ref{fig:complex}, we report the model parameters and per-epoch training efficiency compared to SOTA models. Specifically, our proposed MacNet has a significantly larger number of parameters (27.50M) compared to SOTA models, whose parameters range from 0.199M (DTFD-MIL) to 4.075M (DGR-MIL). Correspondingly, the training time per epoch of MacNet (145s) is also substantially higher than that of other methods, where the most complex model, DGR-MIL requires 9.35s.
 
Despite the computational cost, our approach yields feature representations that are strongly task-related. This results in superior classification performance and interpretability at the cost of acceptable resources.

\section{Conclusion}
WSI-based pathology grading becomes increasingly important for clinical cancer diagnosis and prognosis. However, existing MIL models suffer from degraded representation due to overrepresentation of weakly relevant NTIs and irrelevant BGIs. Therefore, we proposed a novel clustering-based end-to-end learning framework. Extensive experiments demonstrate that our proposed MacNet exhibits superior classification performance. MacNet also exhibits strong interpretability ability in feature representations. Both its active and passive interpretability show high consistency with the ground truth annotations provided by pathologists, which offers a new paradigm for lesion diagnosis, localization and segmentation. Furthermore, we demonstrate that end-to-end learning can achieve better performance than two-stage approaches. It actually does not require demanding computational resources. This work represents a pioneering exploration in end-to-end WSI-based pathology analysis, laying a foundation for more intelligent, interpretable, and unified diagnostic frameworks in computational pathology.

In the future, we will introduce the concept of causal inference to achieve unguided clustering. Also, we will consider incorporating pseudo-label learning guided by reinforcement learning to enhance the alignment between instances and pathological grading.

\appendix
\section{Appendix}
\subsection{Innovated derivation of kernel $\mathbf{G}$}
In original derivation, to obtain the expression of $\Phi(t)$, we need to compute \cite{gong2012geodesic}:
\begin{equation}\label{app:eq1}
        \mathbf{P}_i^T\mathbf{P}_j=\mathbf{V}_1\Gamma\ \mathbf{V}^T, \quad \mathbf{R}_i^T\mathbf{P}_j=-\mathbf{V}_2\Sigma\mathbf{V}^T\
\end{equation}
where $\mathbf{R}_i\in \mathbb{R}^{n\times(n-d)}$ denote the orthogonal complement to $\mathbf{P}_i$. $\Gamma$ and $\Sigma$ are $d\times d$ diagonal matrices. This is a generalized SVD problem numerically implemented in Numpy package. Therefore, We solved \eqref{app:eq1} to retain the tensor in PyTorch computation graph.

Note that $\mathbf{A}=\mathbf{P}_i^T\mathbf{P}_j \in \mathbb{R}^{d\times d}$ and $\mathbf{B}=\mathbf{R}_i^T\mathbf{P}_j \in \mathbb{R}^{(n-d))\times d}$. $\mathbf{A}$ is the product of two numerically column orthogonal matrices (PCA subspaces). As a result, $\mathbf{A}$ can be approximately regarded as orthogonal matrix, i.e. a non-singular matrix. In this case, we can implement this special case of GSVD in PyTorch easily; see the following equations.

Given $\mathbf{A} \in \mathbb{R}^{d \times d}, \mathbf{B} \in \mathbb{R}^{(n-d) \times d}, \text{with } \mathbf{A} \text{ invertible}$. Take the singular value decomposition of $\mathbf{B}\mathbf{A}^{-1} \in \mathbb{R}^{(n-d)) \times d}$:
\[
\mathbf{B}\mathbf{A}^{-1} = -\mathbf{V}_2 \mathbf{S} \mathbf{V}_1^\top
\]
Define the generalized singular values with $\mathbf{\Gamma}^2 + \mathbf{\Sigma}^2 = \mathbf{I}$:
\[
\Gamma_i = \frac{1}{\sqrt{1 + s_i^2}}, \quad \Sigma_i = \frac{s_i}{\sqrt{1 + s_i^2}}
\]
and construct diagonal matrices:
\[
\mathbf{\Gamma} = \mathrm{diag}(\Gamma_1, \dots, \Gamma_d), \quad \mathbf{\Sigma} = \mathrm{diag}(\Sigma_1, \dots, \Sigma_d)
\]
Then, define $\mathbf{V} = \mathbf{A}^T \mathbf{V}_1 \mathbf{\Gamma}^{-1}$.
We obtain the generalized singular value decomposition:
\[
\boxed{
\mathbf{A} = \mathbf{V}_1 \mathbf{\Gamma} \mathbf{V}^\top, \quad 
\mathbf{B} = -\mathbf{V}_2 \mathbf{\Sigma} \mathbf{V}^\top
}
\]

\bibliographystyle{IEEEtran}
\bibliography{IEEEabrv,my_ref}

\end{document}